\documentclass[letterpaper]{article} 
\usepackage{aaai2026}
\nocopyright

\usepackage{times}  
\usepackage{helvet}  
\usepackage{courier}  
\usepackage[hyphens]{url}  
\usepackage{graphicx} 
\usepackage{natbib}  
\usepackage{caption} 
\usepackage{algorithm}
\usepackage{amsmath,amssymb,amsfonts}
\usepackage{algpseudocode}  
\usepackage{booktabs}

\newcommand{\mathbbm}[1]{\mathbf{#1}}
\newcommand{\ind}{\mathbf{1}}

\usepackage{newfloat}
\usepackage{listings}
\DeclareCaptionStyle{ruled}{labelfont=normalfont,labelsep=colon,strut=off} 
\floatstyle{ruled}
\newfloat{listing}{tb}{lst}{}
\floatname{listing}{Listing}
\title{ARISE: Adaptive Agentic Reasoning with Image-grounded Self-Evaluation for Interpretable IBD Assessment}
\author{
    Pronoma Banerjee\textsuperscript{\rm 1},
    Anuva Shah\textsuperscript{\rm 1},
    Jason Wu\textsuperscript{\rm 1},
    Md. Masudur Rahman\textsuperscript{\rm 1},
    Sanjay Mohanty\textsuperscript{\rm 2}, 
    Satya Kurada\textsuperscript{\rm 2}, 
    and Juan P. Wachs\textsuperscript{\rm 1}
}
\affiliations{
    \textsuperscript{\rm 1} Purdue University, West Lafayette, IN 47907, USA\\
    \textsuperscript{\rm 2} Indiana University School of Medicine, Indianapolis, IN 46202, USA

    jpwachs@purdue.edu
}

\usepackage{bibentry}

\begin{document}

\maketitle

\begin{abstract}
Inflammatory bowel disease (IBD) requires frequent imaging-based assessment, yet interpretation of modalities such as wireless capsule endoscopy (WCE) and intestinal ultrasound remains heavily dependent on specialist expertise. Vision-Language Models (VLMs) demonstrate significant potential in multimodal medical image analysis, but their clinical adoption is hindered by their insufficient domain-specific reasoning, susceptibility to hallucination, scarcity of good quality training data in fine-grained diagnostics and limited interpretability. We introduce ARISE (Adaptive Agentic Reasoning with Image-grounded Self-Evaluation), an autonomous planning framework that models few-shot medical image understanding as a sequential agentic workflow. ARISE structures agent execution into a transparent 5-stage workflow: hypothesis generation, image-grounded evidence summarization, evidence-conditioned refinement, symbolic verification, and final diagnosis. We apply ARISE to IBD assessment across two independent patient cohorts: wireless capsule endoscopy (WCE) images for Crohn's disease and B-Mode ultrasound data for Ulcerative Colitis. ARISE consistently improves diagnostic performance over baseline VLMs while exposing where reasoning succeeds or fails, providing a more interpretable basis for clinical decision support and realistic deployment.
\end{abstract}


\section{Introduction}

Inflammatory bowel disease (IBD), including Crohn’s disease (CD) and ulcerative colitis (UC), is a chronic relapsing condition that requires repeated assessment of disease activity to prevent progressive bowel damage and other complications \cite{ng2017worldwide, turner2021stride2, torres2017crohns}. Two imaging modalities are central to this monitoring workflow. Wireless capsule endoscopy (WCE) \cite{iddan2000wireless} provides direct visualization of small-bowel mucosal abnormalities in Crohn’s disease, while intestinal ultrasound (IUS) \cite{maaser2020trustuc} offers a radiation-free means of assessing bowel-wall features such as colon wall thickening and loss of stratification, both established markers of activity in ulcerative colitis \cite{bots2018ultrasound}. Despite their clinical value, both modalities share a critical bottleneck: accurate interpretation depends heavily on scarce, subspecialty-trained experts \cite{vallee2020crohnipi, leenings2025ctrus}. In resource-limited and non-tertiary settings, this dependence delays diagnosis and monitoring precisely where timely intervention matters most \cite{kaplan2021four}.

Vision-language models (VLMs) offer a promising way to support medical image interpretation without task-specific training \cite{radford2021clip, alayrac2022flamingo, liu2023llava, openai2023gpt4}. Yet their translation into IBD assessment is hindered by four well-documented obstacles \cite{hager2024_llm_clinical_decision}. First, general-purpose VLMs exhibit a domain-reasoning gap: fine-grained IBD diagnosis requires integrating subtle, co-occurring visual cues over disease progression rather than recognizing salient objects, and these cues are underrepresented in general training distributions \cite{du2025_mki_prompt_tuning,zhang2024disease}. Second, VLMs remain prone to hallucination and to internally contradictory rationales, an unacceptable failure mode in a clinical setting \cite{ji2023hallucination, li2025treble, madaan2023selfrefine}. Third, high-quality annotated IBD imaging data are scarce, ruling out conventional fine-tuning in most practical deployments \cite{shakeri2024fewshot, vallee2020crohnipi}. 
Finally, even accurate VLMs act as black boxes, providing no way to distinguish sound visual reasoning from chance or spurious shortcut \cite{Byun2025TestTimeScaling, byun2026medicalvqacalibration}; inference-time methods such as test-time prompt tuning and retrieval-augmented in-context learning avoid retraining, but do not necessarily produce reasoning that can be inspected or verified\cite{shu2022tpt, ram2023incontextralm, zhan2026retrieval, xia2025mmedrag}. 

Neurosymbolic methods that translate natural-language problems into symbolic form for solver-backed verification improve logical faithfulness, but they have been developed for text-only reasoning rather than image-grounded medical diagnosis, where hypotheses must be tied to observable visual findings \cite{pan2023logiclm, olausson2023linc, garcez2023neurosymbolicai}. Moreover, most medical-agent evaluations emphasize final-task accuracy, making it difficult to identify where failures arise within a multi-stage reasoning process \cite{liu2023mmbench, fu2025mmebenchmark, hager2024_llm_clinical_decision}. As recent workflow-aware benchmarks have argued, final-output metrics alone cannot reveal where or why an agent fails \cite{schmidgall2024agentclinic, jiang2025medagentbench}. In short, no existing few-shot adaptation approach produces a reasoning state that is simultaneously grounded in the image evidence, verified for internal consistency, and open to expert inspection before deployment.

We introduce ARISE (Adaptive Agentic Reasoning with Image-grounded Self-Evaluation), an autonomous framework that adapts a frozen VLM to few-shot IBD assessment by constructing, grounding, verifying, and refining an explicit set of diagnostic hypotheses, and that evaluates both its intermediate reasoning stages and its final diagnosis. ARISE follows five stages: (S1) generation of diagnostic hypotheses from task and clinical knowledge, (S2) image-grounded summarization of findings from a small support set, (S3) evidence-conditioned refinement of those hypotheses, (S4) symbolic verification and repair using first-order logic and an SMT solver, and (S5) final diagnosis of unseen images. This design makes transparency and correctness mutually reinforcing, and constrains hallucination without fine-tuning. Each stage produces an inspectable artifact and an associated score, allowing ARISE to evaluate both the quality of its own reasoning process and the accuracy of the resulting diagnosis evaluated in two settings: Wireless Capsule Endoscopy (WCE) for Crohn’s disease and B-mode ultrasound for ulcerative colitis.

Our contributions are fourfold: First, we introduce a few-shot adaptation strategy in which domain knowledge is represented as image-grounded, iteratively refined hypotheses rather than learned parameters. Second, we extend solver-backed neurosymbolic reasoning to image-grounded medical diagnosis by verifying and repairing VLM-generated hypotheses before prediction. Third, we introduce stage-level evaluation that separates agentic process quality from task performance, revealing where reasoning failures occur rather than reducing evaluation to a final label. Finally, we demonstrate the framework across independent WCE \cite{vallee2020crohnipi} and B-mode ultrasound \cite{leenings2025ctrus} cohorts and multiple VLM families, showing that structured, verifiable reasoning can improve diagnostic performance across modalities and model backbones.

\section{Methodology}
\label{sec:method}

\begin{figure*}
    \centering
    \includegraphics[width=0.8\linewidth]{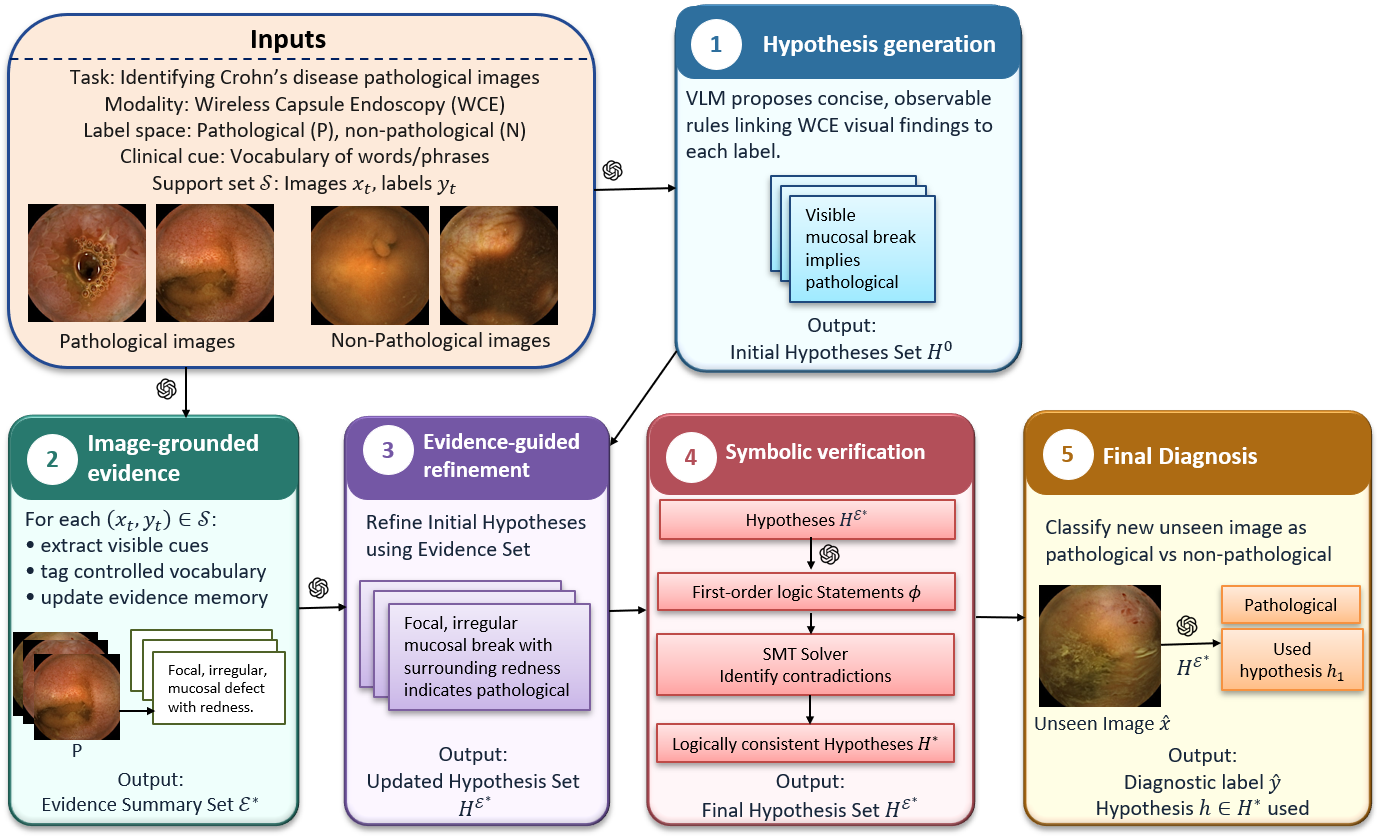}
    \caption{An overview of the ARISE workflow for the Crohn's disease WCE Pathological vs Non-Pathological image classification. The workflow starts with initial hypotheses generation and a parallel summarization of visual cues from select examples. The initial hypotheses are first updated using the summary of visual cues, followed by a logical consistency check via an SMT solver. The final hypotheses are used for the final diagnosis stage. }
    \label{fig:pipeline}
\end{figure*}
 
We describe \textbf{ARISE} (\emph{Adaptive Agentic Reasoning with Image-grounded Self-Evaluation}), an autonomous agent that adapts a frozen vision-language model (VLM) to few-shot inflammatory bowel disease (IBD) assessment by executing a transparent, five-stage reasoning workflow. Unlike a single opaque prompt, ARISE carries an explicit workflow \emph{state} across stages, invokes an external symbolic tool, and emits an
inspectable artifact at every stage that receives its own \emph{score}. No VLM parameters are updated: adaptation is stored entirely as an auditable hypothesis state. This eliminates the dependency on massive datasets, bypassing the conventional fine-tuning bottleneck that typically hinders medical AI deployment. We summarize a run as an \emph{agentic} (process) score over the five workflow stages plus a \emph{task} (outcome) score, and combine them into a single overall score.
 
\subsection{Problem Formulation}
\label{sec:problem}
 
Let $\mathcal{D}=\{(x_i,y_i)\}_{i=1}^{N}$ be an IBD imaging
dataset, where $x_i$ is an input image (a wireless capsule endoscopy (WCE) frame for the Crohn's cohort or a B-mode ultrasound frame for the ulcerative colitis (UC) cohort) and $y_i\in\mathcal{Y}$ is a diagnostic label. We operate in the few-shot regime: a small labeled support set $\mathcal{S}=\{(x_j,y_j)\}_{j=1}^{m}$ is available for adaptation, and a disjoint
query set $\mathcal{Q}$ is held out. Each task carries a task description $T$
(modality, label space $\mathcal{Y}$, acquisition context, requested output) and a clinical cue description $C$ enumerating expert knowledge about the visual findings associated with each label.
 
The \textbf{ARISE agent} is the tuple $(M_\theta,\pi,V,\Gamma)$: $M_\theta$ is
the frozen VLM; $\pi$ is a deterministic controller that sequences the stages and triggers repair loops; $V$ is a symbolic satisfiability-modulo-theories (SMT) verifier (Z3); and $\Gamma$ is a prompt-based parser that maps each
natural-language hypothesis to a typed first-order-logic (FOL) premise. The agent maintains a state
\begin{equation}
\sigma^{(t)}=\bigl(H^{(t)},\ \Phi^{(t)},\ \mathcal{E}^{(t)},\ \mathbf{z}^{(t)}\bigr),
\end{equation}
comprising the hypothesis set $H^{(t)}$, its premises
$\Phi^{(t)}=\Gamma(H^{(t)})$, an evidence memory $\mathcal{E}^{(t)}$, and the
vector of stage scores $\mathbf{z}^{(t)}=(s_1,\dots,s_5)$. The exact scores are described in details in Appendix A. The goal is to construct a verified hypothesis state $H^{\ast}$ that yields an accurate query prediction $\hat{y}=f_{\mathrm{task}}(x,H^{\ast},\mathcal{E}^{\ast},M_\theta)$
while producing an inspectable, individually-scored reasoning trace.
 
\subsection{Stage S1: Hypothesis Generation}
\label{sec:w1}
 
ARISE composes a planning prompt from task and clinical knowledge,
$p_0=\mathrm{Plan}(T,C,\mathcal{Y})$, and the VLM generates concise, checkable
hypotheses  
\begin{equation}
H^{(0)}=\{h_1,\dots,h_K\}=M_\theta(p_0;\ \tau,p_{\mathrm{top}}),
\end{equation}
where the sampling temperature $\tau$ and nucleus sampling parameter $p_{\text{top}}$ can be varied to encourage hypothesis diversity. For example, in the WCE six-lesion task, $p_0$ may ask: ``Given Crohn's disease capsule endoscopy images and the label set $\{\mathrm{AU}, \mathrm{E}, \mathrm{O}, \mathrm{S}, \mathrm{U3-10}, \mathrm{U{>}10}\}$, generate concise visual hypotheses linking observable mucosal findings to each label. Consider mucosal breaks, white ulcer base, red halo, diffuse redness, enlarged villi, and luminal narrowing. Avoid claims that cannot be checked from the image.''

 
\subsection{Stage S2: Image-Grounded Evidence Summarization}
\label{sec:w2}
 
ARISE initializes an empty evidence memory $\mathcal{E}^{(0)}=\varnothing$ and picks one example image at a time, sequentially. At step $t$ it selects $(x_t,y_t)$ and
asks the VLM to report only the \emph{observable} evidence in $x_t$ supporting
$y_t$ under the current hypotheses:
\begin{equation}
e_t=M_\theta\bigl(\mathrm{Observe}(x_t,y_t,p_0)\bigr).
\end{equation}
The output is \emph{structured}: $e_t$ tags each cited finding against a
per-cohort controlled vocabulary $\mathcal{V}$ (for endoscopy, this will be mucosal break, white ulcer base, red halo, diffuse erythema, enlarged villi, luminal narrowing, etc.). This makes scoring a set comparison rather than text parsing, which is simpler to evaluate. The generated visual summary is appended as
$\mathcal{E}^{(t)}=\mathcal{E}^{(t-1)}\cup\{e_t\}$. The final evidence set is denoted as $\mathcal{E}^*$.
The grounding score \(s_2\) measures agreement between model-identified findings and expert-annotated findings, emphasizing visual fidelity over unsupported inference.

\subsection{Stage S3: Evidence-Conditioned Refinement}
\label{sec:w4}
 
Given the accumulated evidence, the agent proposes a refined set
$H^{(\mathcal{E}^*)}=M_\theta(\mathrm{Refine}(p_0,H^{(0)},\mathcal{E}^*))$. The refinement score \(s_3\) measures whether the refined hypothesis set improves downstream prediction performance relative to the initial hypotheses on a held-out validation subset. 

\begin{figure}

    \centering
    \includegraphics[width=1\linewidth]{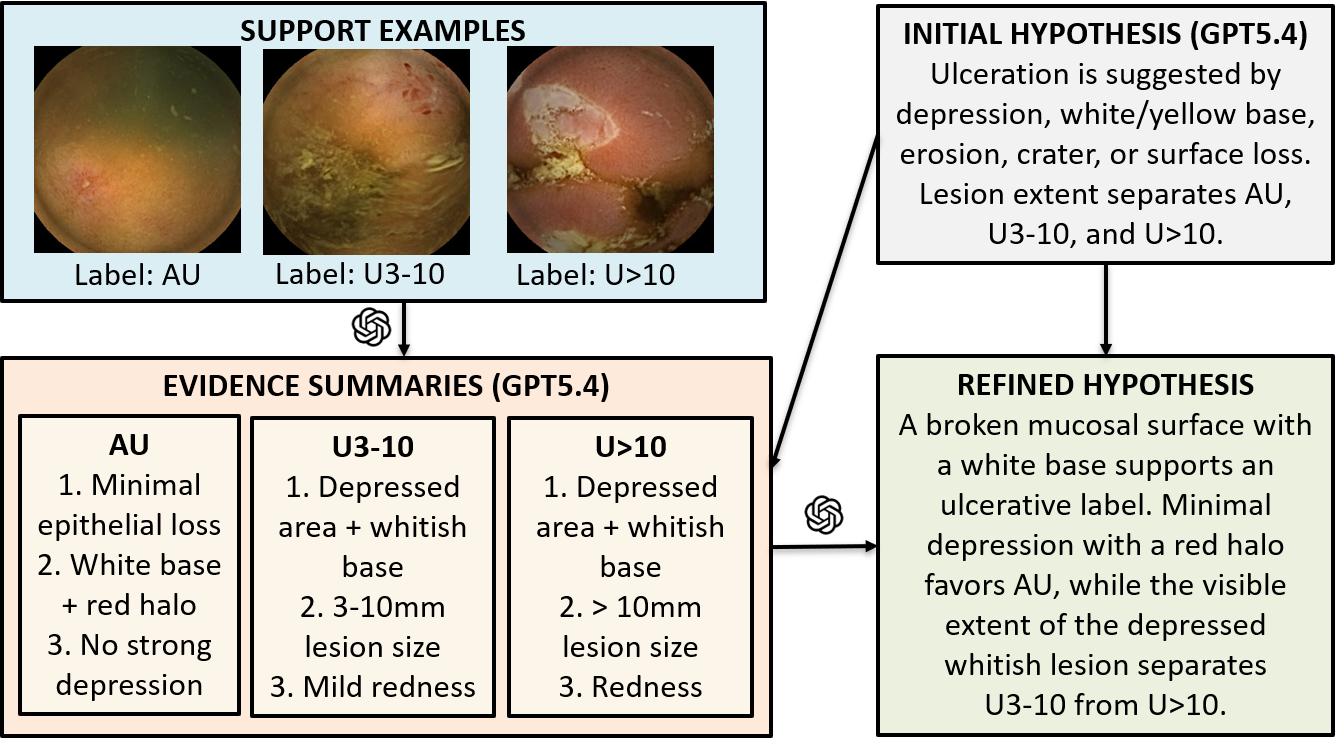}
    \caption{Example of stages S2 and S3 on the Crohn's disease Ulceration Diagnosis task with GPT-5.4}
    \label{fig:refinementexample}
\end{figure}

Stages S2 and S3 are jointly described in Figure \ref{fig:refinementexample}. 

\subsection{Stage S4: Symbolic Verification}
\label{sec:w3}
 
The generated hypotheses $H^{(\mathcal{E}^*)}$ are converted from natural language to first order logic statements $\Phi$ using an LLM parser $\Gamma$,  $\Phi = \Gamma(H^{(\mathcal{E}^*)})$. For example, the natural language statement "A mucosal break with a white base indicates aphthoid ulceration" is translated to the form $(\text{MucosalBreak}(x) \wedge \text{WhiteBase}(x)) \rightarrow \text{AU}$. These are then passed into a Satisfiability Modulo Theories Solver.
\begin{equation}
V(\Phi)=
\begin{cases}
1,& \text{if } \Phi \text{ is satisfiable},\\
0,& \text{otherwise.}
\end{cases}
\end{equation}
If $V(\Phi)=0$, $V$ returns the minimal conflicting set $b$ and the LLM repairs the hypotheses,
$H^{*}=M_\theta(\mathrm{Repair}(p_0,H^{(\mathcal{E}^*)},b))$, giving the final hypotheses set $H^{*}$ to be used for diagnosis.  

The verification score \(s_4\) measures the fidelity of the natural-language-to-logic translation, evaluated through bidirectional entailment between the original and reconstructed hypotheses. This is done using a natural language inference model, details are provided in the Appendix A.

\subsection{Stage S5: Final Diagnosis}
\label{sec:w5}
 
With the verified state $H^{\ast}$, the model performs a downstream classification task, classifying an input $\hat{x}$ with a corresponding label $\hat{y}$. A few examples of pictures and labels are added for in-context learning. For each classification, we also output the hypothesis ${h}$ it uses for the classification. The stage score \(s_5\) measures whether the prediction belongs to the valid label space and whether the model grounds its prediction in the verified hypothesis set.

The task score for the entire agentic system is measured as the average of the classification accuracy, F1-score, precision, and recall on $\mathcal{Q}$.

\subsection{Workflow-Aware Scoring: Agentic $+$ Task}
\label{sec:scoring}
 
The five workflow stages are aggregated into a single \emph{agentic} (process) score, and the outcome is the \emph{task} score:
\begin{align}
\textsc{Agentic} &= \sum_{k=1}^{5} w_k\,s_k,\qquad \sum_{k=1}^{5} w_k=1, \\
\textsc{Overall} &= \tfrac{1}{2}\,\textsc{Agentic}+\tfrac{1}{2}\,\textsc{Task}.
\end{align}
Weights emphasize the trust-critical stages, with grounding ($s_2$) assigned the highest weight and symbolic verification and refinement the second highest; our default is $\mathbf{w}=(w_1,\dots,w_5)=(0.20,0.30,0.20,0.20,0.10)$.

\section{Experiments}
\subsection{Datasets and Tasks}
 
ARISE is evaluated across two IBD imaging domains: Crohn’s disease (CD) using wireless capsule endoscopy (WCE) and ulcerative colitis (UC) using B-mode intestinal ultrasound. Detailed dataset composition, support/few-shot examples/test splits, and task definitions are provided in Appendix B.
 
For Crohn's disease, we use the CrohnIPI dataset \cite{vallee2020crohnipi}, a wireless capsule endoscopy (WCE) collection of 3,498 RGB images independently reviewed by multiple clinical experts. We evaluate three tasks of increasing diagnostic granularity: (1) Pathological vs non-pathological image classification, (2) ulceration diagnosis (ulceration vs. no ulceration), and (3) six-lesion classification for pathological images, with labels being aphthoid ulceration, stenosis, edema, erythema, ulceration between $3$ and $10$ mm, and ulceration \(>10\) mm.

For Ulcerative Colitis, ultrasound data is taken from the C-TRUS dataset \cite{leenings2025ctrus}, containing a total of 800 ultrasound images collected from 13 patients. We evaluate colon-wall stratification status, classifying whether bowel-wall layering is preserved or lost, a clinically relevant marker of intestinal inflammation.
 
\subsection{Models and Prompting}
 
We evaluate ARISE using instruction-following VLMs from three model families: GPT, Gemini, and Qwen. We test GPT-5.4 \cite{afridi2026gpt}, GPT-4o-mini \cite{openai2023gpt4}, Gemini-2.5 \cite{comanici2025gemini25pushingfrontier}, and Qwen2-VL \cite{wang2024qwen2vlenhancingvisionlanguagemodels}, with Qwen2-VL serving as an open-weight VLM baseline. 
All models are evaluated on identical query splits. Automatic hypothesis generation and logical premise construction use a reasoning-oriented language model, while symbolic contradiction detection is performed using the Z3 SMT solver~\cite{demoura2008z3}.

We perform 2 major experiments. 1. We compare the diagnostic accuracy, F1 score, precision, and recall of the base VLM and the VLM with the final generated hypotheses (task score after stage 5). 2. We then compare the agentic scores for every stage across models, and the overall task score for every model, on the 6-lesion classification task. This would tell us which stages are strongest and weakest, and which models tend to outperform the others in both workflow and task-level performance. Crucially, these evaluations jointly establish the clinical value of verifiable reasoning over standard prompting, while opening the black box to reveal exactly how process-level reasoning quality drives final diagnostic accuracy.

\section{Results}

\subsection{Diagnostic Performance: Base VLM vs. ARISE}

Table \ref{tab:combined_accuracy} compares base-model accuracy with the same VLMs operating within the ARISE workflow across the three Crohn’s disease WCE tasks. ARISE improves accuracy for every model, with the largest accuracy gain observed as the task complexity increases (6-lesion classification being the hardest) and for the weaker baseline models. Qwen2-VL and Gemini-2.5 close much of the gap to the stronger GPT baselines once wrapped in the ARISE workflow, suggesting that the framework's gains come from structuring and verifying reasoning rather than from any one backbone's raw visual capability. Detailed task-level results for this dataset, including accuracy, F1 score, precision, and recall for all model–task combinations, are provided in the Appendix.

For the ulcerative-colitis cohort, Table \ref{tab:colitis} reports the corresponding accuracy, F1, precision, and recall for the B-mode ultrasound colon-wall-stratification task (preserved vs. lost). The ARISE framework results in a consistent improvement over all baseline models even when diagnosing from noisy B-mode ultrasound images, showing that the performance improvement generalizes across different tasks and modalities. 

\begin{table}
\centering
\caption{Base and ARISE classification accuracy across the three Crohn's disease WCE tasks. Task 1: Pathological vs non-pathological image classification, Task 2: Ulceration diagnosis, Task 3: 6-lesion classification}
\label{tab:combined_accuracy}
\begin{tabular}{p{0.38\linewidth}ccc}
\toprule
\textbf{Model}  &
\textbf{Task 1} &
\textbf{Task 2} &
\textbf{Task 3} \\
\midrule

GPT-5.4 (Base)  & 0.69 & 0.66 & 0.32 \\
GPT-5.4 (ARISE) & \textbf{0.88} & \textbf{0.83} & \textbf{0.79} \\

GPT-4o-mini (Base)  & 0.61 & 0.61 & 0.29 \\
GPT-4o-mini (ARISE) & \textbf{0.88} & {0.81} & {0.74} \\

Gemini-2.5 (Base)  & 0.66 & 0.33 & 0.37 \\
Gemini-2.5 (ARISE) & {0.85} & {0.71} & {0.58} \\

Qwen2-VL (Base)  & 0.42 & 0.66 & 0.22 \\
Qwen2-VL (ARISE) & {0.83} & {0.74} & {0.64} \\
\bottomrule
\end{tabular}
\end{table}

\begin{table}
\caption{Colon wall stratification status classification}
\label{tab:colitis}
\centering
\footnotesize
\begin{tabular}{p{0.38\linewidth}cccc}
\toprule
Model & Accuracy & F1 & Precision & Recall \\
\midrule
GPT-5.4 (Base) & 0.60 & 0.66 & 0.69 & 0.63\\{
GPT}-5.4 (ARISE)& \textbf{0.87} & \textbf{0.78} & \textbf{0.83} & 0.74\\
GPT-4o-mini (Base) & 0.59 & 0.58 & 0.58 & 0.47\\
GPT-4o-mini (ARISE) & 0.77 & {0.75} & {0.76} & 0.74 \\
Gemini-2.5 (Base) & 0.26 & 0.37 & 0.63 & 0.22 \\
Gemini-2.5 (ARISE) & 0.71 & 0.78 & 0.76 & 0.72 \\
Qwen2-VL (Base) & 0.66 & 0.64 & 0.63 & 0.64 \\
Qwen2-VL (ARISE) & 0.74 & {0.74} & 0.75 & \textbf{0.76} \\

\bottomrule
\end{tabular}
\end{table}

\subsection{Stage-Level Reasoning Quality}

\begin{figure*}
    \centering
    \includegraphics[width=0.95\linewidth]{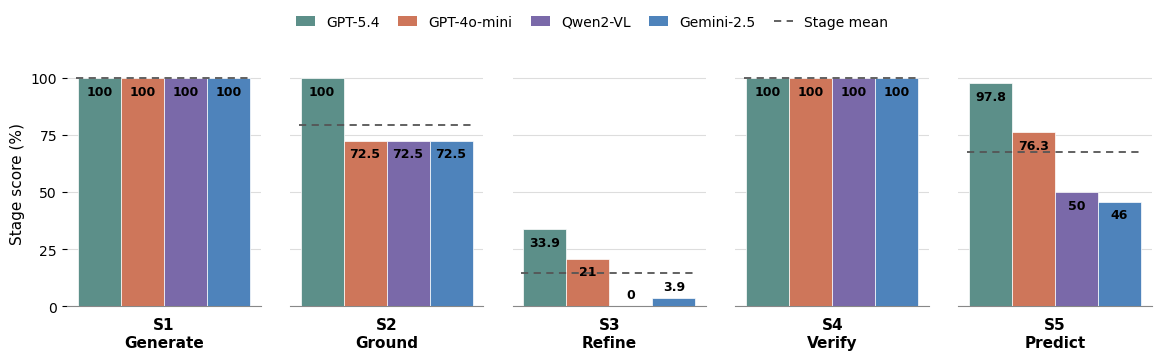}
    \caption{Stage-wise agentic scores across models on the Crohn's disease 6-lesion classification task for the different workflow stages: hypothesis generation (S1), visual grounding (S2), refinement (S3), symbolic verification (S4), and final diagnosis (S5). Rule-based stages are consistently strong, while evidence-guided refinement shows the greatest difficulty and model variation.}
    \label{fig:stagewiseagenticscores}
\end{figure*}

Figure \ref{fig:stagewiseagenticscores} reports the five individual stage scores (s1 through s5) for all four models on the six-lesion classification task, together with the mean across models at each stage.
Hypothesis generation (S1) and symbolic verification (S4) are consistently strong across models, while evidence-guided refinement (S3) is the most challenging stage and exhibits the greatest model variation. The contrast between the two kinds of stages points to a broader pattern: agents are consistently better at rule-based steps than at visually-grounded refinement, where they must revisit the image evidence itself and use it to revise their own hypotheses rather than follow a fixed procedure. The strongest models also perform better during refinement and final diagnosis, indicating that visually grounded adaptation is a key factor separating stronger from weaker agents.

\subsection{Agentic and Task-level scores across models}

\begin{figure}
    \centering
    \includegraphics[width=0.9\linewidth]{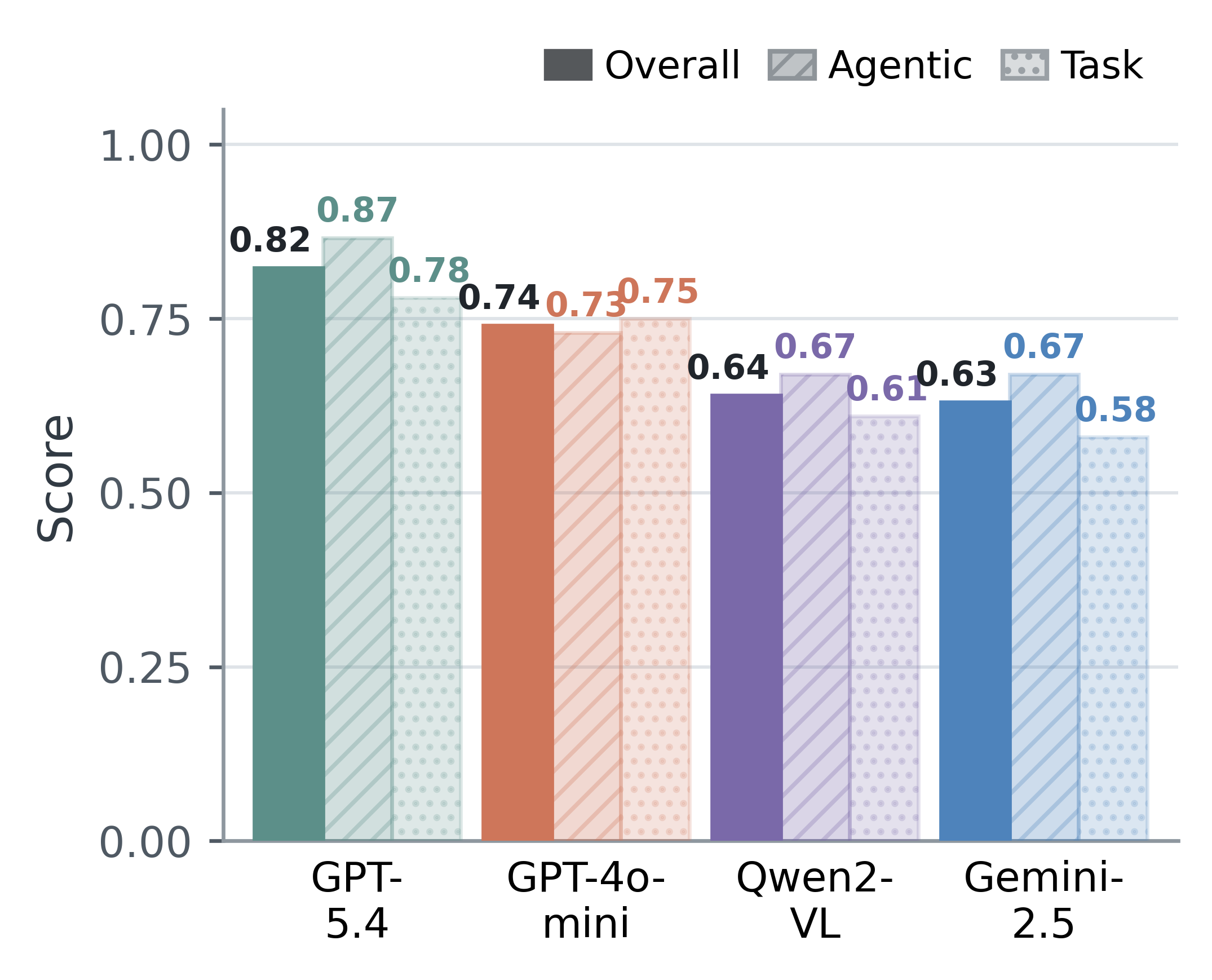}
    \caption{Overall, agentic, and task scores for the 4 evaluated agents on the Crohn's disease 6-lesion classification task. }
    \label{fig:overallagenttaskscores}
\end{figure}

We examine how each model's intermediate reasoning quality (the agentic score) relates to its final classification performance (the task score) on the six-lesion classification task. Figure \ref{fig:overallagenttaskscores} reports the overall, agentic, and task scores for all four models, ranked by overall score.

Across models, stronger agentic performance generally corresponds to stronger diagnostic performance, but the relationship is not one-to-one. The rule-based stages, such as hypothesis generation and symbolic verification, are consistently strong across all models, whereas the largest differences emerge during evidence-guided refinement and final diagnosis, where the model must use visual evidence to revise and apply its hypotheses. Hence weaker models can maintain relatively high agentic scores by successfully completing structured stages such as hypothesis generation and verification, while still struggling to translate that reasoning into the correct final classification. The resulting process–outcome gap highlights the value of ARISE’s stage-level evaluation, which reveals where reasoning quality stops carrying through to diagnostic performance.

\subsection{Qualitative Visual Grounding}

\begin{figure}
    \centering
    \includegraphics[width=1\linewidth]{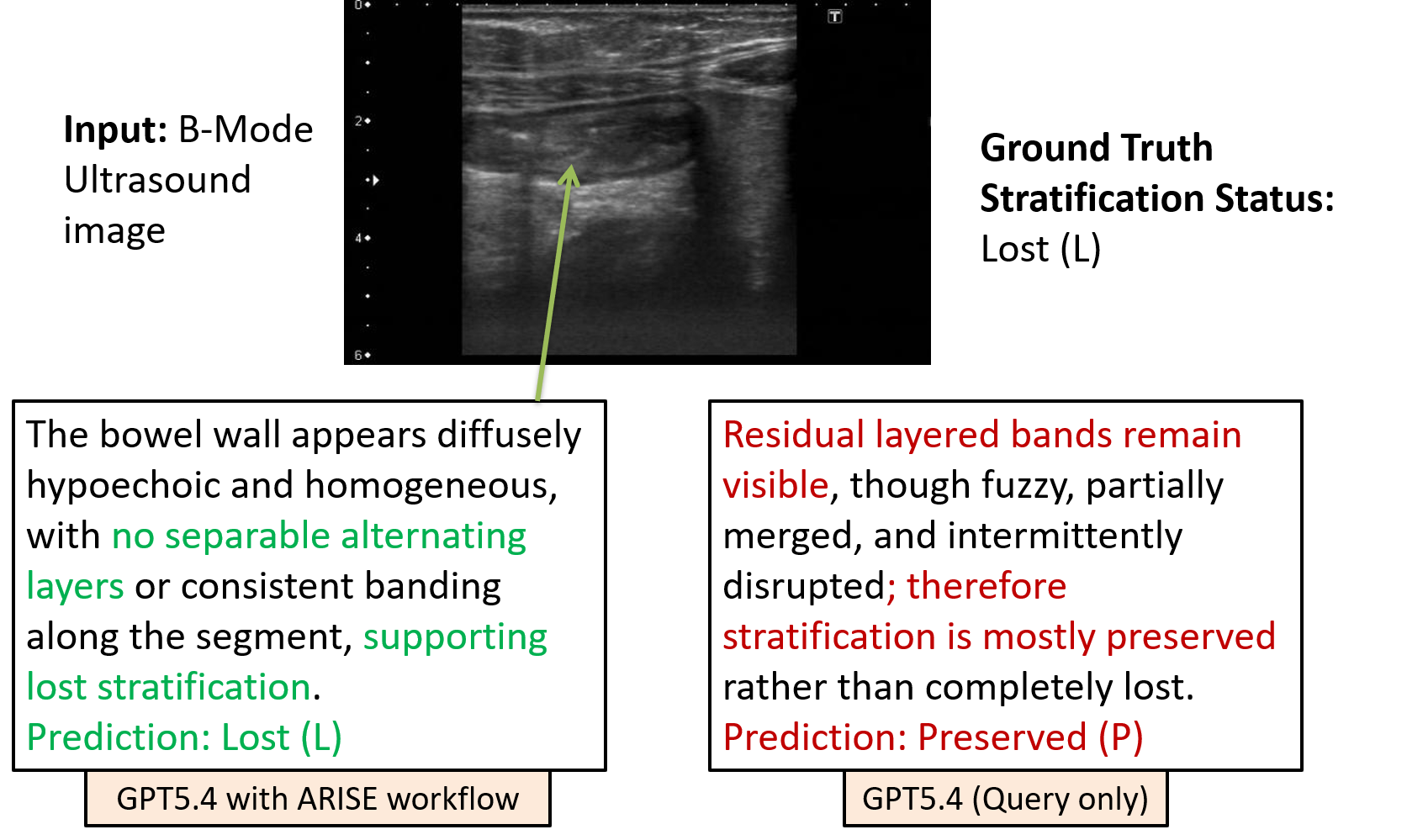}
    \caption{Qualitative example of improvement of visual grounding with ARISE. The base VLM produces ambiguous visual cues, ultimately leading to the wrong classification. ARISE improves visual grounding and ultimately yields the correct classification. }
    \label{fig:qualitative}
\end{figure}

To illustrate the mechanism behind these quantitative gains, Figure \ref{fig:qualitative} presents a representative B-mode ultrasound case from the colon-wall-stratification task with ground-truth label "lost." Queried directly, GPT-5.4 hedges between competing visual interpretations and concludes an incorrect diagnosis. Under the ARISE workflow, the same underlying model grounds its answer in concrete, vocabulary-anchored observations ("the bowel wall appears diffusely hypoechoic and homogeneous, with no separable alternating layers or consistent banding along the segment") and correctly predicts "lost." This case is typical of the pattern underlying the quantitative improvements described earlier: ARISE's evidence-summarization (S2) and symbolic-verification (S4) stages guide the model away from ambiguous, hedged visual claims and toward grounded and verifiable ones, which is where the framework's accuracy gains originate.

\section{Path to Deployment}

\begin{figure}
\centering
\includegraphics[width=1\linewidth]{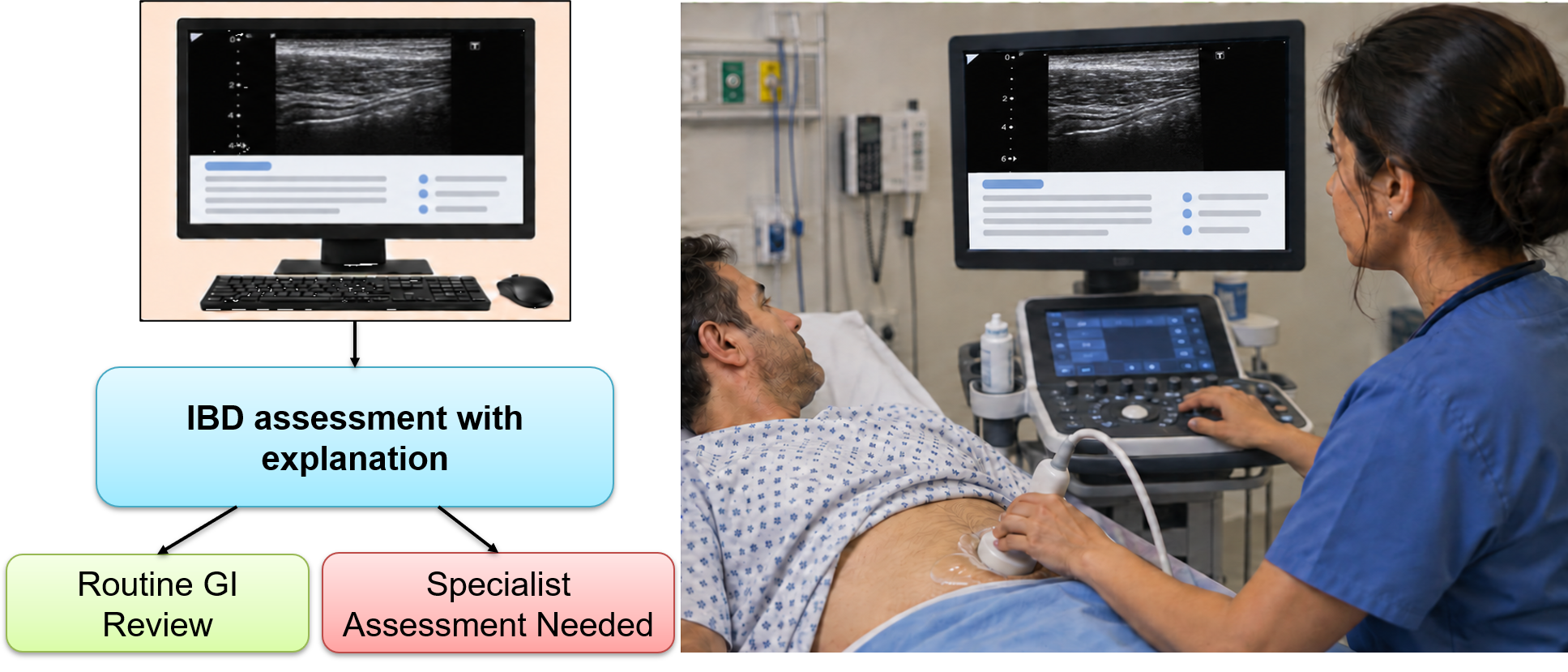}
\caption{Illustration of an AI-enabled decision-support interface for explainable IBD assessment using intestinal ultrasound, designed to make model outputs accessible and interpretable to nurses.}
\label{fig:deployment}
\end{figure}

ARISE is designed as a clinician-in-the-loop decision-support system rather than an autonomous diagnostic tool, emphasizing interpretability, error detection, and compatibility with existing IBD imaging workflows. The current framework has been evaluated across two complementary IBD imaging settings: 3,498 wireless capsule endoscopy (WCE) images from the publicly available CrohnIPI dataset \cite{vallee2020crohnipi} and B-mode intestinal ultrasound data obtained from the C-TRUS dataset \cite{leenings2025ctrus} and annotated by an expert at the Indiana University (IU) School of Medicine. Clinical experts informed the visual cues, decision boundaries, and interpretation guidelines used by ARISE. Unlike black-box deep-learning based models, ARISE exposes the visual findings, diagnostic hypotheses, symbolic verification results, refinement steps, and final recommendation, allowing clinicians to inspect or override reasoning before acting on the output.

The next deployment stage is retrospective clinical validation using anonymized IU School of Medicine data from approximately 40 adults with Crohn's disease, collected by a gastrointestinal ultrasound expert certified by the International Bowel Ultrasound Group (IBUS), an international organization supporting standardized training, research, and clinical adoption of intestinal ultrasound for IBD care. Expert annotations of bowel-wall stratification and clinical assessments will serve as reference standards for evaluating ARISE predictions. Stage-level errors will be analyzed across disease severity, patient variation, image quality, and acquisition hardware.

For patients with anonymized, clinically indicated endoscopy data, ARISE will also estimate the Simple Endoscopic Score for Crohn's Disease (SES-CD) \cite{DAPERNO2004505}, from AI-identified ulcer size, ulcerated surface, and luminal narrowing across the terminal ileum and four colonic segments. This will be compared to the SES-CD score assessments by experts. 

Following retrospective validation, ARISE will be incorporated into a software-based clinical interface capable of processing endoscopy and B-mode ultrasound data streams. In an initial shadow-mode study, ARISE outputs will be compared with specialist interpretation without influencing patient care. A subsequent clinician-in-the-loop pilot will allow specialists to accept, question, or override intermediate reasoning and final recommendations, providing direct evidence of where the system's explanations are clinically useful or insufficient.

Finally, nurses and generalist clinicians will participate in a small case study evaluating whether ARISE's visual-cue explanations improve non-expert interpretation and diagnostic accuracy relative to unaided assessment (Figure \ref{fig:deployment}). Successful prospective evaluation would support testing ARISE across multiple hospitals and integrating it with existing clinical imaging and reporting systems. This staged progression, from retrospective validation, to silent real-time testing, to clinician-assisted evaluation and multi-site deployment, provides a practical path toward safe, interpretable use of ARISE in IBD assessment. 

\section{Conclusion}

We presented ARISE, a five-stage agentic framework that adapts a frozen vision-language model to few-shot IBD assessment by generating, grounding, verifying, and refining diagnostic hypotheses without any fine-tuning. Across two independent cohorts (WCE for Crohn's disease and B-mode ultrasound for ulcerative colitis) and three VLM families, ARISE improved accuracy, F1, precision, and recall over baselines on every model–task combination, with the largest gains on the hardest task and weakest baseline model, while its stage-level scoring exposed cases where strong intermediate reasoning did not fully carry through to final accuracy, highlighting a gap a single outcome metric would hide. Beyond raw accuracy, ARISE's key contribution is auditability: every reasoning stage is stored as an inspectable artifact rather than collapsed into an opaque prediction, directly addressing the domain-reasoning gap, hallucination risk, data scarcity, and black-box behavior limiting clinical VLM adoption. Future work extends this toward improving the model with the agentic scores and deployment in real-world clinical settings.

\bibliography{references}

\appendix

\setcounter{table}{0}  \renewcommand{\thetable}{A\arabic{table}}
\setcounter{equation}{0}  \renewcommand{\theequation}{A\arabic{equation}}

\section{A. Stage Scoring Definitions and Equations}

\paragraph{Notation.}
Let $\mathcal{Y}$ denote the task label space, $H^{(0)}$ the initial hypothesis set, $\mathcal{E}^*$ the accumulated image-grounded evidence, $H^*$ the final verified hypothesis set, $\mathcal{V}$ the controlled clinical vocabulary, $\Gamma$ the natural-language-to-first-order-logic parser, and $\Phi=\Gamma(H)$ the corresponding logical premises. $\hat{y}$ denotes the predicted label and $\hat{h}$ the hypothesis cited in support of that prediction.

\paragraph{Score $s_1$.} S1 is scored by a three-item rubric averaged to
$[0,1]$:
\begin{align}
s_1 &= \tfrac{1}{2}\Bigl(\underbrace{\mathbbm{1}\!\left[H^{(0)}\neq\varnothing\right]}_{\text{non-empty}}
+ \underbrace{\frac{\sum_{y\in \mathcal{Y}} \mathbbm{1} \left[{\exists h\in H^{(0)}, x \rightarrow^{h} y\ }\right]}{|\mathcal{Y}|}}_{\text{per-label coverage}} \Bigr)
\end{align}

where the first term indicates that the hypotheses set generated has at least 1 hypothesis, and the 2nd term indicates that there is at least one hypothesis per label. The best agentic score ($s_1 = 1$) is obtained when both conditions hold. 

\paragraph{Score $s_2$.} Let $\mathcal{F}(e_t)$ be the findings cited by the
agent and $\mathcal{F}^{\star}(x_t)$ the label-relevant findings marked present
by expert annotation of the (few) support images. We report precision and
recall separately and combine them, weighting fidelity above completeness since
a fabricated finding is the clinical dealbreaker:
\begin{align}
\mathrm{Prec} &= \frac{|\mathcal{F}(e_t)\cap\mathcal{F}^{\star}(x_t)|}{|\mathcal{F}(e_t)|}, \quad
\mathrm{Rec} = \frac{|\mathcal{F}(e_t)\cap\mathcal{F}^{\star}(x_t)|}{|\mathcal{F}^{\star}(x_t)|}, \\
\mathrm{Obs} &= \frac{|\{c\in e_t:\, c\in\mathcal{V}\}|}{|e_t|}, \\
s_2 &= 0.5 \mathrm{Prec}+0.4\mathrm{Rec}+0.1\mathrm{Obs}
\end{align}
where $\mathrm{Obs}$ (observability) is the fraction of claims referencing a
concrete vocabulary finding rather than an abstract inference. 

\paragraph{Score $s_3$.} Refinement is scored by the improvement in prediction accuracy of downstream task before and after the refinement step, when compared against the ground truth label, for a smaller validation set of 50 observations. 
\begin{align}
s_3 = \text{Acc}_{H^{(\mathcal{E}^*)}} - \text{Acc}_{H^{(0)}}
\end{align}

\paragraph{Score $s_4$.} We score the translation fidelity of the LLM parser using a Natural Language Inference (NLI) model (DeBERTa-v3-large) ~\cite{he2021deberta}. First, we compute the natural language hypothesis back from the first order logic:

\begin{equation}
\tilde{H^{*}} = \Gamma^{-1}(\Phi)
\end{equation}

We then test whether $H^{*}$ and $\hat{H^{*}}$ mutually entail one another (each statement implies the other). The NLI gives us the entailment probability (probability of one statement implying the other) and we get the score as the min of 2 probabilities. 
\begin{align}
s_4 = \min\left[P_{\mathrm{e}}\!\left(H^* \rightarrow \tilde{H}^*\right),
P_{\mathrm{e}}\!\left(\tilde{H}^* \rightarrow H^* \right)\right]    
\end{align}

This ensures that the 2 statements are equivalent and the first order logic statement was correct. 

\paragraph{S5: Final diagnosis.}
The process score checks whether the prediction lies in the permitted label space and whether the model cites a hypothesis from the verified set as support:
\begin{equation}
s_5 = \frac{1}{2}\left[
\ind(\hat{y}\in\mathcal{Y})+\ind(\hat{h}\in H^*)
\right].
\label{eq:s5}
\end{equation}

\paragraph{Task score.}
Diagnostic outcome quality is measured separately from process quality as the mean of accuracy, F1 score, precision, and recall on the held-out query set:
\begin{equation}
\mathrm{TASK}=\frac{1}{4}\left[
\mathrm{Acc}+\mathrm{F1}+\mathrm{Prec}+\mathrm{Rec}
\right].
\label{eq:task}
\end{equation}

\paragraph{Workflow-level aggregation.}
The five stage scores are combined using the weights reported in the main manuscript, with the greatest weight assigned to image grounding:
\begin{align}
\mathrm{AGENTIC} &= \sum_{k=1}^{5}w_k s_k,\\
\mathrm{OVERALL} &= \frac{1}{2}\mathrm{AGENTIC}+\frac{1}{2}\mathrm{TASK}.
\end{align}

where $(w_1,\ldots,w_5) = (0.20,0.30,0.20,0.20,0.10)$

\section*{B. Datasets and experimental protocol}

\paragraph{Crohn's disease (Wireless Capsule Endoscopy)}
The CrohnIPI dataset \cite{vallee2020crohnipi} contains 3,498 expert-reviewed WCE RGB images. The manuscript reports 60.55\% normal and 38.85\% pathological images. Pathological images include erythema, edema, aphthoid ulceration, ulceration between 3 and 10mm, ulceration greater than 10mm, and stenosis; some images contain co-occurring labels.

\paragraph{Ulcerative colitis (B-Mode Ultrasound)}
The C-TRUS dataset \cite{leenings2025ctrus} contains 800 B-mode ultrasound images from 13 patients. The study filters 275 good-quality images for the colon-wall stratification experiment. These raw datapoints were annotated for colon-wall stratification classification, and segmentation of the colon wall, by an expert at IU School of Medicine. 

The datasets, tasks and number of support and in-context examples and number of test samples are mentioned in Table \ref{tab:splits}

\begin{table*}[t]
\centering
\small
\caption{Dataset and task splits used in the experiments. AU: aphthoid ulceration; S: stenosis; O: edema; E: erythema; U3-10: ulceration between 3 and 10 mm; U$>$10: ulceration $>10$~mm.}
\label{tab:splits}
\begin{tabular}{p{0.26\textwidth}p{0.17\textwidth}p{0.28\textwidth}ccc}
\toprule
Task & Modality & Output labels & Support & ICL & Test \\
\midrule
Normal vs. pathological classification & WCE & Normal; pathological & 7 & 18 & 1434 \\
Six-lesion classification & WCE  & AU; S; O; E; U3--10; U$>$10 & 7 & 12 & 535 \\
Ulceration diagnosis & WCE  & Ulceration; no ulceration & 7 & 12 & 535 \\
Colon-wall stratification status & B-mode ultrasound & Preserved; lost & 4 & 10 & 211 \\
\bottomrule
\end{tabular}
\end{table*}

\subsection*{Experimental Protocol}
\begin{enumerate}
    \item \textbf{Base VLM evaluation:} Each backbone is first evaluated in a query-only setting on the held-out test split.
    \item \textbf{ARISE evaluation:} The same backbone is evaluated on the same query split after the five-stage ARISE workflow constructs the verified hypothesis state $H^*$.
    \item \textbf{Task-level comparison:} Accuracy, F1 score, precision, and recall are reported for Base and ARISE on all four clinical tasks.
    \item \textbf{Stage-level analysis:} Scores S1--S5 are compared across the four VLMs on the Crohn's disease six-lesion classification task. S3 is computed on a separate 50-example validation subset.
    \item \textbf{Process vs. outcome analysis:} The weighted AGENTIC score is compared with the TASK score to identify cases in which a well-formed reasoning process does not fully translate to correct final classification.
\end{enumerate}

\section{C. Results for Crohn's disease }

Tables \ref{tab:wce_normal_path}, \ref{tab:ulceration} and \ref{tab:lesion_classification} below report the full task-level metrics for each backbone in the query-only Base setting and within ARISE.

\begin{table} [t]
\caption{Pathological vs Non-Pathological WCE image}
\label{tab:wce_normal_path}
\centering
\footnotesize
\begin{tabular}{p{0.35\linewidth}cccc}
\toprule
Model & Accuracy & F1 & Precision & Recall \\
\midrule
GPT-5.4 (Base)  & 0.69 & 0.65 & 0.59 & 0.72 \\
GPT-5.4 (ARISE) & \textbf{0.88} & 0.86 & 0.85 & \textbf{0.89} \\
GPT-4o-mini (Base) )  & 0.61 & 0.73 & 0.60 & 0.77 \\
GPT-4o-mini (ARISE) & \textbf{0.88} & \textbf{0.88} & \textbf{0.89} & 0.88 \\
Gemini-2.5 (Base)  & 0.66 & 0.59 & 0.77 & 0.62 \\
Gemini-2.5 (ARISE) & 0.85 & 0.78 & 0.82 & 0.76 \\
Qwen2-VL (Base)  & 0.42 & 0.29 & 0.21 & 0.50 \\
Qwen2-VL (ARISE) & 0.83 & 0.83 & 0.88 & 0.83 \\

\bottomrule
\end{tabular}
\end{table}

\begin{table}[t]
\caption{WCE - Diagnosing ulceration}
\label{tab:ulceration}
\centering
\footnotesize
\begin{tabular}{p{0.38\linewidth}cccc}
\toprule
Model & Accuracy & F1 & Precision & Recall \\
\midrule
GPT-5.4 (Base) & 0.66 & 0.76 & 0.79 & 0.73\\
GPT-5.4 (ARISE) & \textbf{0.83} & 0.78 & 0.83 & 0.74\\
GPT-4o-mini (Base)  & 0.61 & 0.67 & 0.86 & 0.59\\
GPT-4o-mini (ARISE) & 0.81 & \textbf{0.84} & \textbf{0.88} & 0.81 \\
Gemini-2.5 (Base)  & 0.33 & 0.18 & 0.78 & 0.10 \\
Gemini-2.5 (ARISE) & 0.71 & 0.78 & 0.86 & 0.72 \\
Qwen2-VL (Base)  & 0.66 & 0.77 & 0.76 & 0.79 \\
Qwen2-VL (ARISE) & 0.74 & \textbf{0.84} & 0.75 & \textbf{0.96} \\
\bottomrule
\end{tabular}
\end{table}

\begin{table}
\caption{6 lesion classification from Crohn's disease WCE}
\label{tab:lesion_classification}
\centering
\footnotesize
\begin{tabular}{p{0.38\linewidth}cccc}
\toprule
Model & Accuracy & F1 & Precision & Recall \\
\midrule
GPT-5.4 (Base) & 0.32 & 0.30 & 0.35 & 0.32 \\
GPT-5.4 (ARISE) & \textbf{0.79} & \textbf{0.76} & \textbf{0.81} & \textbf{0.75}  \\
GPT-4o-mini (Base) & 0.29 & 0.34 & 0.32 & 0.26 \\
GPT-4o-mini(ARISE) & 0.74 & 0.75 & 0.77 & 0.74 \\
Gemini-2.5 (Base) & 0.37 & 0.35 & 0.47 & 0.37 \\
Gemini-2.5(ARISE) & 0.58 & 0.53 & 0.58 & 0.63 \\
Qwen2-VL (Base) & 0.22 & 0.11 & 0.07 & 0.22 \\
Qwen2-VL (ARISE) & 0.64 & 0.58 & 0.58 & 0.64\\

\bottomrule
\end{tabular}
\end{table}

\section*{D. Prompt Templates and Implementation}


\subsection*{Prompt A: Normal vs. Pathological WCE Classification}
The binary prompt assigns each WCE frame to exactly one of two labels, \texttt{N} (normal/non-pathological) or \texttt{P} (pathological). The prompt defines pathological findings to include aphthoid ulceration, ulceration, erythema, edema, stenosis, focal mucosal breaks, depressed lesions, white/yellow ulcer bases, red halos, abnormal redness, swelling, or luminal narrowing. The decision rules instruct the model to choose \texttt{N} only when the mucosa appears continuous and no focal abnormality is visible, and to choose \texttt{P} whenever a definite abnormality is present. In the baseline setting, no refined hypotheses or in-context examples are provided. In the ARISE setting, the refined hypothesis list is inserted under a \emph{Supporting rules} field and the model is asked to return the predicted class together with the ID of the supporting hypothesis when available.

\begin{quote}\small
\textbf{Prompt A template:}\\
You are a medical assistant specialized in wireless capsule endoscopy image classification.\\
Classify each WCE image into exactly one binary label: \\
\texttt{N} = normal-appearing / non-pathological frame.\\
\texttt{P} = pathological / abnormal frame.\\[0.25em]
\textbf{Supporting rules:} [refined hypotheses, ARISE setting only]\\[0.25em]
\textbf{Decision rules:} choose \texttt{N} only for normal, continuous mucosa without focal abnormality; choose \texttt{P} if any visible abnormality is present; do not output a lesion subtype.
\end{quote}

\subsection*{Prompt B: Six-Lesion WCE Classification}
Prompt B is applied only to frames already considered pathological and assigns exactly one lesion label: aphthoid ulceration (\texttt{AU}), erythema (\texttt{E}), edema (\texttt{O}), stenosis (\texttt{S}), ulceration 3--10 mm (\texttt{U3-10}), or ulceration $>$10 mm (\texttt{U>10}). The prompt includes a short description of each class and an ulcer-first decision hierarchy: visible mucosal break, depression, erosion, crater, or white/yellow ulcer base directs the model toward an ulcer subtype; intact-surface redness favors erythema; swollen/enlarged villi with a continuous surface favors edema; and dominant luminal narrowing favors stenosis. The baseline Prompt B uses a fixed set of short class-discrimination rules. In the ARISE setting, these fixed rules are replaced by the refined hypothesis state loaded from disk.

\begin{quote}\small
\textbf{Prompt B template:}\\
You are a medical assistant specialized in wireless capsule endoscopy image classification.\\
This image is already considered pathological. Do not choose \texttt{N}.\\
Classify the pathological subtype into exactly one of: \\
\texttt{AU, E, O, S, U3-10, U>10}.\\[0.25em]
\textbf{Supporting rules:} [fixed short rules for baseline; refined hypotheses for ARISE]\\[0.25em]
\textbf{Decision hierarchy:} ulcerative surface loss $\rightarrow$ ulcer subtype; intact-surface redness $\rightarrow$ \texttt{E}; swollen/enlarged villi with continuous surface $\rightarrow$ \texttt{O}; dominant luminal narrowing $\rightarrow$ \texttt{S}.
\end{quote}

\subsection*{Support and In-Context Examples}
A set of support examples are selected for each task for Stage S2. Seprately, ARISE runs use manually selected labeled image examples as multimodal in-context demonstrations. The demonstrations contain the example image and its class label; the explanatory text stored with each example is used during the final diagnostic evaluation. For Prompt A, GPT and Gemini use four ICL examples, while Qwen2-VL-2B uses two. For Prompt B, the GPT notebook permits up to 12 examples, Gemini uses four, and Qwen2-VL-2B uses two. Qwen uses fewer examples to reduce GPU memory consumption on a Colab T4.


\end{document}